\documentclass[11pt]{article}
\usepackage{acl2015}

\usepackage{graphics}
\usepackage{epsfig}
\usepackage{times}
\usepackage{amsmath}

\usepackage[usenames,dvipsnames]{color}
\usepackage{tikz}
\usepackage{tikz-qtree}
\usepackage{tikz-dependency}

\usepackage[linesnumbered,vlined,ruled]{algorithm2e}
\usepackage{graphicx}
\usepackage{wrapfig}
\usepackage{dblfloatfix} 
\usepackage{caption}
\usepackage{subcaption}
\usepackage{fancybox}

\usepackage{rotating}

\usepackage{url}
\usepackage{latexsym}

\usepackage{amsfonts}
\usepackage{multirow}
\usepackage{verbatim}

\usepackage{array}

\usepackage{amssymb}

\usepackage{caption}
\title{OmniGraph: Rich Representation and Graph Kernel Learning}

\author{Boyi Xie \\
  Department of Computer Science \\
  Columbia University \\
  New York, NY, USA \\
  {\tt xie@cs.columbia.edu} \\\And
  Rebecca J. Passonneau \\
  Center for Computational Learning Systems \\
  Columbia University \\
  New York, NY, USA \\
  {\tt becky@ccls.columbia.edu} \\}

\date{}

\begin{document}
\maketitle
\begin{abstract}
OmniGraph, a novel representation to support a range of NLP classification tasks, integrates lexical items, syntactic dependencies and frame semantic parses into graphs. Feature engineering is folded into the learning through convolution graph kernel learning to explore different extents of the graph. A high-dimensional space of features includes individual nodes to complex networks. In experiments on a text-forecasting problem that predicts stock price change from news for company mentions, OmniGraph beats several benchmarks based on bag-of-words, syntactic dependencies, and semantic trees. The highly expressive features OmniGraph discovers provide insights into the semantics across distinct market sectors. To demonstrate the method's generality, we also report its high performance results on a fine-grained sentiment corpus. 
\end{abstract}

\section{Introduction}

\begin{figure}[t]
\begin{center}
\includegraphics[width=.4\textwidth]{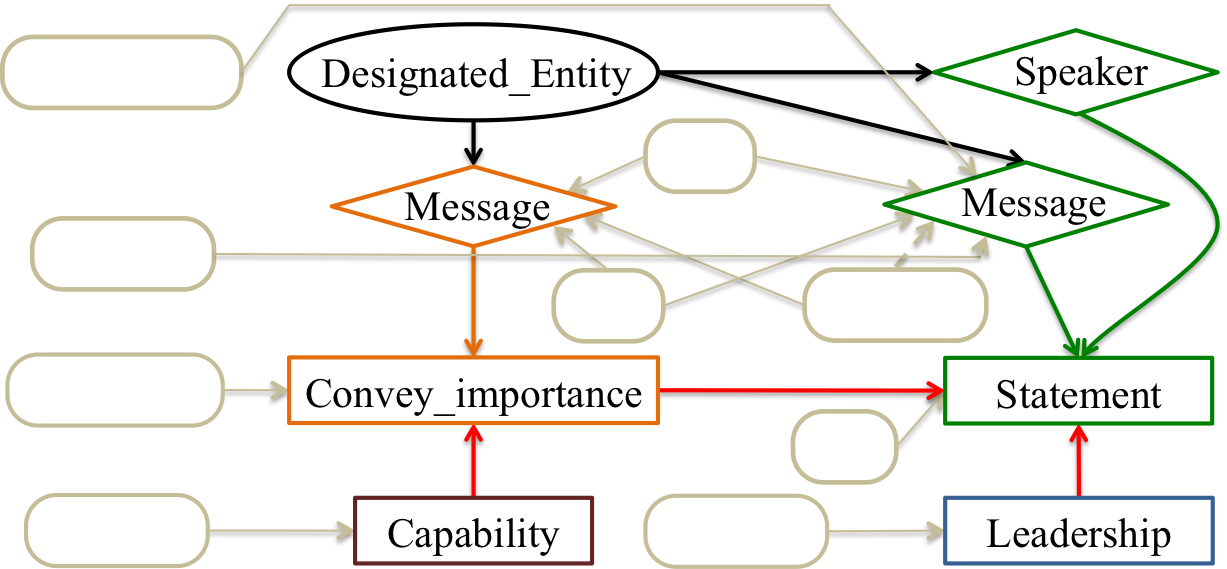}
\end{center}
\vspace{-.05in}
\small{\textit{``This milestone \textcolor{orange}{highlighted} the {\bf Boeing} KC-767's \textcolor{brown}{ability} to perform refueling operations under all lighting conditions,'' \textcolor{OliveGreen}{said} George Hildebrand, {\bf Boeing} Japan \textcolor{blue}{program manager}.}}\newline
\vspace{-.1in}
\caption{\small{The graph for the sentence appears with nodes and edges greyed out, apart from the colored subgraph, which is a predictive OminGraph feature consisting of frame names (rectangles), semantic roles (diamonds), and semantic and syntactic relations (arrows). It predicts a positive price change for Boeing (boldface).}}
  \label{fig:capability-feature}
\end{figure}

For diverse NLP classification tasks, such as sentiment and opinion mining, or text-forecasting, in which text documents are used to make predictions about measurable phenomena in the real world~\cite{Kogan2009}, there is a need to generalize over words while simultaneously capturing relational and structural information. Feature engineering for NLP learning tasks can be labor-intensive. We propose OmniGraph, a novel representation that supports a continuum of features from lexical items, to syntactic dependencies, to frame semantic features. Figure~\ref{fig:capability-feature} illustrates a sentence, the structure of its graph, and a predictive subgraph feature our method discovers that captures semantic and syntactic dependencies (arrows), semantic role information for syntactic arguments (diamonds), and generalizations over lexical items (semantic frame names, shown as rectangles). For machine learning with OmniGraph, we use graph kernels that allow the user to control how much of the graph is explored for similarity computation. We test this approach on an extremely challenging text-forecasting problem: a polarity classification task to predict the direction of price change for publicly traded companies based on news. We also report results on an entity-driven fine-grained sentiment corpus. 

The ability to exploit deep semantic information in text, e.g. to distinguish the depicted scenarios and semantic roles of the entity mentions, motivates our study. We hypothesize that a general and uniform representation of linguistic information that combines multiple levels, such as semantic frames and roles, syntactic dependency structure and lexical items, can support challenging classification tasks for NLP problems. Consider the following three sentences from financial news articles. 

\begin{small}
- \textit{``The accreditation renewal also \textcolor{orange}{\underline{underscores}} the \textcolor{brown}{\underline{quality}} of our work with \textbf{Humana} members, customers, clients, payors and health care providers by confirming our compliance with national standards for PBM services," \textcolor{OliveGreen}{\underline{said}} William Fleming, \textcolor{blue}{\underline{vice president}} of \textbf{Humana Pharmacy Solutions}.} 

- \textit{``The testing program \textcolor{orange}{\underline{highlighted}} the \textcolor{brown}{\underline{abilities}} of the Navy, \textbf{Raytheon Missile Systems} and NASA to effectively partner on this complicated program and deliver what would have been previously unobtainable data,'' \textcolor{OliveGreen}{\underline{said}} Don Nickison, \textcolor{blue}{\underline{chief}} of the NASA Ames Wind Tunnel operations division.} 

- \textit{``The initiation of a dividend and the renewed share repurchase authorization \textcolor{orange}{\underline{underscore}} the board  and management's confidence in \textbf{Symantec}'s long-term business outlook and \textcolor{brown}{\underline{ability}} to generate significant free cash flow on a consistent basis," \textcolor{OliveGreen}{\underline{said}} \textbf{Symantec}'s \textcolor{blue}{\underline{executive vice president and} \underline{chief financial officer}}, James Beer.} 

\end{small}

\noindent
The sentences all describe a scenario in which a company executive makes a positive statement about the company's capabilities. Of note, the stock price of the three companies (in boldface) went up the next day. Four semantic frames from FrameNet~\cite{baker98}, a linguistic resource that exemplifies Fillmore's frame semantics \cite{fillmore76}, capture the commonality of a {\it statement} (green) from an organization {\it leader} (blue) that {\it conveys the importance} (orange) of {\it capabilities} (brown). Further, within each sentence the frames have the same syntactic dependencies in the three sentences. The feature in Figure~\ref{fig:capability-feature} captures the common meaning of these sentences, and is predictive in three distinct market sectors: industrials, health care and information technology.

\section{Related Work}
\label{sec:related}

Much recent work on the kinds of NLP classification tasks our experiments address, text-forecasting and fine-grained sentiment, builds on linguistically informed features or knowledge. \newcite{kim06} introduce fine-grained opinion mining, using semantic role labeling to mine triples of the source, target and content of opinions applied to online news. To similarly mine opinion triples, \newcite{sayeed12} depend more on syntax, using a suffix-tree data structure to represent syntactic relationships. Instead of feature engineering, \newcite{yogatama&smith14}, develop structured regularization for \textsc{bow} based on parse trees, topics and hierarchical word clusters to improve \textsc{bow} for 3 classification tasks: topic, sentiment, and text-driven forecasting. Another approach to forecasting from text \cite{joshi-EtAl:2010:NAACLHLT} combines \textsc{bow} and the names of dependency relations to engineer features for predicting movie revenue from reviews. They devote considerable effort to feature engineering, while our approach folds feature engineering into the learning. 

OmniGraph feature engineering is handled automatically by convolution graph kernels. Convolution kernels have been used in NLP to exploit structured information using trees for parsing and tagging \cite{collins01}, text categorization \cite{lodhi2002}, and question answering \cite{zhang&lee03,suzuki2003,moschitti06}. To learn social networks, Agarwal et al. \shortcite{agarwalEtAl14} use partial tree kernels on a representation with frame semantic information \cite{fillmore76}.  The tree representation in \newcite{xie2013} also incorporates frame semantics, and uses subtree and subset tree kernels for the same forecasting task we pursue. In contrast to their task-specific representations, our more general OmniGraph can be used for many tasks. Rather than having to choose among many tree kernels, the graph kernels we use allow users to specify the size of the graph neighborhoods to explore. 

Studies of the effect of financial news on the market \cite{gerberEtAl09,gentzkow&shapiro10,engelberg&parsons11} have been increasingly important since Tetlock \shortcite{Tetlock2007} investigated the role of media in the stock market. As mentioned in \newcite{Wong14}, a better solution to the problem can help gain more insights to the long-lasting question in finance about how financial markets react to news~\cite{Fama98marketefficiency,Chan2003}. In general, the work in NLP that uses news to predict price does well if it achieves better than 50\% accuracy \cite{Lee2014,barhaim-EtAl2011,Creamer2013,xie2013}. \newcite{Wong14} report that even ``textbook models'' that uses time series data have less than 51.5\% prediction accuracy. Unlike many other domains, however, a higher than random accuracy can have great value in a high-volume trading strategy. Work in NLP and related areas \cite{devitt-ahmad2007,schumaker2012,Feldman2011,zhang2010trading} often treats stock price prediction from news as a sentiment classification problem. \newcite{xie2013} point out that this is consistent with the direction component of the three-part ADS model \cite{Rydberg2003}. Their model was shown better than \textsc{bow} alone on three market sectors, but there was no comparison to the majority baseline. In contrast, our OmniGraph outperforms the baseline in seven out of eight market sectors and beats \textsc{bow} and two other benchmarks.

\begin{figure*}
\begin{small}
\begin{center}
\begin{subfigure}[b]{\textwidth}
\textbf{Sentence:} ``The accreditation renewal underscores the quality of our work with \textbf{Humana} members," said \textbf{Humana}'s president.
\end{subfigure}
~\\
\begin{subfigure}[b]{\textwidth}
\textbf{Frame semantic parse:}\newline \textcolor{OliveGreen}{[}``The accreditation renewal \textcolor{Bittersweet}{[}underscores$\textcolor{Bittersweet}{_{Convey\_importance}}$\textcolor{Bittersweet}{]} \textcolor{Bittersweet}{[}the \textcolor{Brown}{[}quality$\textcolor{Brown}{_{Capability}}$\textcolor{Brown}{]} of our work with \textbf{Humana} members$\textcolor{Bittersweet}{_{Convey\_importance.Message}}$\textcolor{Bittersweet}{]},"$\textcolor{OliveGreen}{_{Stmt.Message}}$\textcolor{OliveGreen}{]} \textcolor{OliveGreen}{[}said$\textcolor{OliveGreen}{_{Statement}}$\textcolor{OliveGreen}{]} \textcolor{OliveGreen}{[}\textbf{Humana}'s \textcolor{blue}{[}president$_{\textcolor{blue}{Leadership}}$\textcolor{blue}{]}   $\textcolor{OliveGreen}{_{Stmt.Speaker}}$\textcolor{OliveGreen}{]}.
\end{subfigure}
\begin{subfigure}[b]{\textwidth}
	\textbf{Dependency parse:}
	\begin{center}
	\begin{dependency}
		\begin{deptext}[column sep=0.001cm]
			``The \& accreditation \& renewal \& \textcolor{Bittersweet}{\underline{underscores}} \& the \& \textcolor{Brown}{\underline{quality}} \& of \& our \& work \& with \& \textbf{Humana} \& members," \& \textcolor{OliveGreen}{\underline{said}} \& \textbf{Humana}'s \& \textcolor{blue}{\underline{president}} \\
		\end{deptext}
		\deproot[edge unit distance=1.5ex]{13}{root}
		\depedge[edge unit distance=0.6ex, edge style={red!60!black,very thick}]{13}{4}{vmod}
		\depedge[edge unit distance=1.8ex]{4}{3}{sub}
		\depedge[edge unit distance=1.8ex, edge style={red!60!black,very thick}]{4}{6}{obj}
		\depedge[edge unit distance=1.8ex]{3}{1}{nmod}
		\depedge[edge unit distance=1.8ex]{3}{2}{nmod}
		\depedge[edge unit distance=1.8ex]{6}{5}{nmod}
		\depedge[edge unit distance=1.8ex]{6}{7}{nmod}
		\depedge[edge unit distance=1.8ex]{7}{9}{pmod}
		\depedge[edge unit distance=1.8ex]{9}{8}{nmod}
		\depedge[edge unit distance=1.8ex]{9}{10}{nmod}
		\depedge[edge unit distance=1.8ex]{10}{12}{pmod}
		\depedge[edge unit distance=1.8ex]{12}{11}{nmod}
		\depedge[edge unit distance=1.8ex, edge style={red!60!black,very thick}]{13}{15}{sub}
		\depedge[edge unit distance=1.8ex]{15}{14}{nmod}
	\end{dependency}
	\label{dep_parse}
	\end{center}
\end{subfigure}
\caption{\small{Example sentence, the frame semantic parse, and the dependency parse. The thick red edges in the dependency parse are dependency relations among the lexical items that trigger frames. Edges corresponding to these dependencies are shown as the red edges in Figure~\ref{fig:semgraph_all}.}}
\label{fig:s1}
\end{center}
\end{small}
\end{figure*}

\begin{figure}[h]
        \centering
        \includegraphics[width=0.5\textwidth]{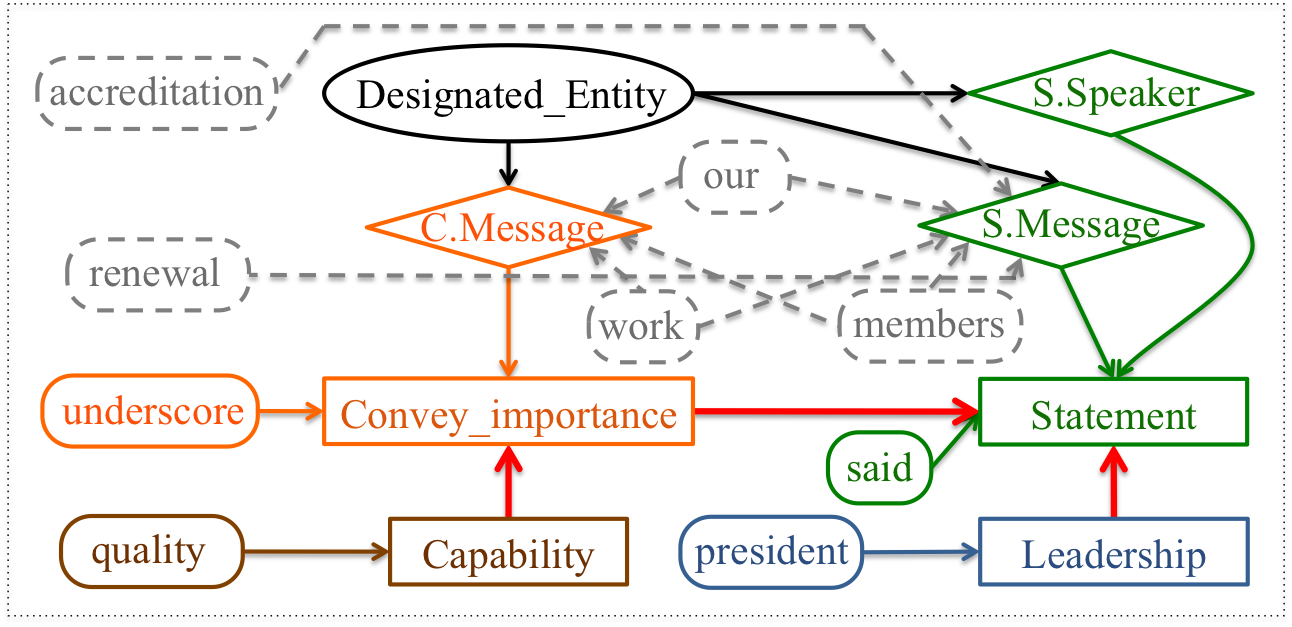} 
        \caption{\small{OmniGraph representation with lexical, dependency, and semantic information for \textit{\textbf{Humana}} in the sentence from Figure~\ref{fig:s1}. For readability, some of the edges are omitted.}} 
	\label{fig:semgraph_all}
\end{figure}

\section{Methods}
\label{sec:methods}

We first introduce our data representation, then describe our learning methods. The nodes in an OmniGraph encode semantic and lexical content, and the edges encode semantic and syntactic dependency relations. In Section~\ref{sec:og} we use an example to describe how to construct a one-sentence OmniGraph. Later we introduce the data instances for our learning task: OmniGraph forests that represent all the sentences in the news that mention a given company on a particular day. As in \newcite{xie2013}, we refer to the company we make predictions about as the designated entity.

\subsection{\textit{OmniGraph} Construction}
\label{sec:og}

To construct a one-sentence OmniGraph, the sentence must first be assigned a frame semantic parse and a syntactic dependency parse. Section \ref{sec:experiments} describes the parsers we use and their performance. We use the example sentence in Figure~\ref{fig:s1} to illustrate how to construct its OmniGraph in Figure~\ref{fig:semgraph_all}. 
 
\noindent \textbf{(1) Create building blocks by converting each semantic frame into a subgraph.} 
In frame-based semantic parsing, the scenarios in a sentence are identified as frames, and each frame is triggered by a frame target: the lexical item that evokes the frame. The frame name and frame target become OmniGraph nodes, along with the frame elements (semantic roles) that have been filled by sentential arguments. In Figure \ref{fig:s1}, nodes with the same color correspond to a frame, its target, and its elements. Four frames have been identified by the parser, and three frame elements: the \textsc{Message} elements of the \textit{Statement} and \textit{Convey\_importance} frames, the \textsc{Speaker} element of the \textit{Statement} frame, and the \textit{Capability} and \textit{Leadership} frames. An edge connects a frame target and the frame it evokes (e.g. \textit{said} and \textit{Statement}), and a frame element and the frame it belongs to (e.g. \textsc{Message} and \textit{Statement}). Figure \ref{fig:s1} shows an actual parse for a sentence in our data; in a correct parse, the phrase \textit{accreditation renewal} would fill the \textsc{Medium} element of the \textit{Convey\_importance} frame.  We achieve good prediction performance despite such inaccuracies.

\noindent \textbf{(2) 
 Add the dependency relations among frames.} Frame semantic parsing identifies individual frames, but not the syntactic dependencies among frames. As shown in the dependency parse in Figure~\ref{fig:s1}, we locate the frame targets, i.e. lexical items that evoke frames, and use the dependency relations among the frame targets to link the frames. In OmniGraph, a dependent node points to the node it depends on, as indicated by the red arrows in Figure~\ref{fig:semgraph_all}. 

\noindent \textbf{(3) Connect the designated entity to its semantic roles.} For the learning task we present here, we make predictions about a designated entity, a publicly traded company. Company mentions are identified using pattern matching. All mentions of the designated entity become nodes, which are linked to the frame elements they fill, or partly fill. For example the designated entity \textbf{Humana} fills the \textsc{Speaker} role of the \textit{Statement} frame, and also occurs in the \textsc{Message} element of the \textit{Statement} frame and the \textsc{Message} element of the \textit{Convey\_importance} frame. It thus has three out-degree edges. 

\noindent \textbf{(4) Connect lexical items to frame elements they help fill.} OmniGraph incorporates lexical information with one node for each lexical item in a constituent that fills a frame element. The nodes connect to the frame elements they fill. Some of these edges are shown as grey dashed nodes and edges in Figure~\ref{fig:semgraph_all}. Some edges are omitted for readability. 

In sum, OmniGraph nodes are: 
1) frame names (boxes),
2) frame targets (rounded boxes), 
3) frame elements (diamonds), 
4) other lexical items (dashed boxes), and
5) designated entities (ellipses).
An edge connects: 
1) a frame target to its frame; 
2) a frame element to the frame it belongs to; 
3) a designated entity to the frame element it fills; 
4) from one frame to another where the target of the first frame is a dependent node in a dependency parse, and the target of the second frame is the dependent's head; 
and 5) a lexical item to the frame element it helps fill. 
Edge directions are exploited by the graph kernels we use.

\subsection{Weisfeiler-Lehman Graph Kernel}
\label{sec:wl}

We selected the Weisfeiler-Lehman (WL) graph kernel \cite{shervashidze2011} for SVM learning because it has a lower computational complexity compared to other graph kernels, and because it can measure similarity between graphs for different neighborhood sizes. At each degree $i$ of neighborhood, all nodes are relabeled with their neighborhoods, then graph similarity is measured. For example, to explore its first degree neighbors, the immediate neighborhood of the \textit{Designated\_Entity} node in Figure \ref{fig:semgraph_all} is used to relabel the node as \textit{\{Designated\_Entity$\to$C.Message, S.Message, S.Speaker\}}. The WL kernel computation is based on the Weisfeiler-Lehman test of isomorphism \cite{weisfeiler1968}, which iteratively augments the node labels by the sorted set of their neighboring node labels, and compresses them into new short labels, called multiset-labels. Through neighbor augmentation, similarity between graphs is iteratively measured using dynamic programming. 

\vspace{-0.1in}
\begin{figure}[ht]
\vspace{.08in}
\includegraphics[width=3in]{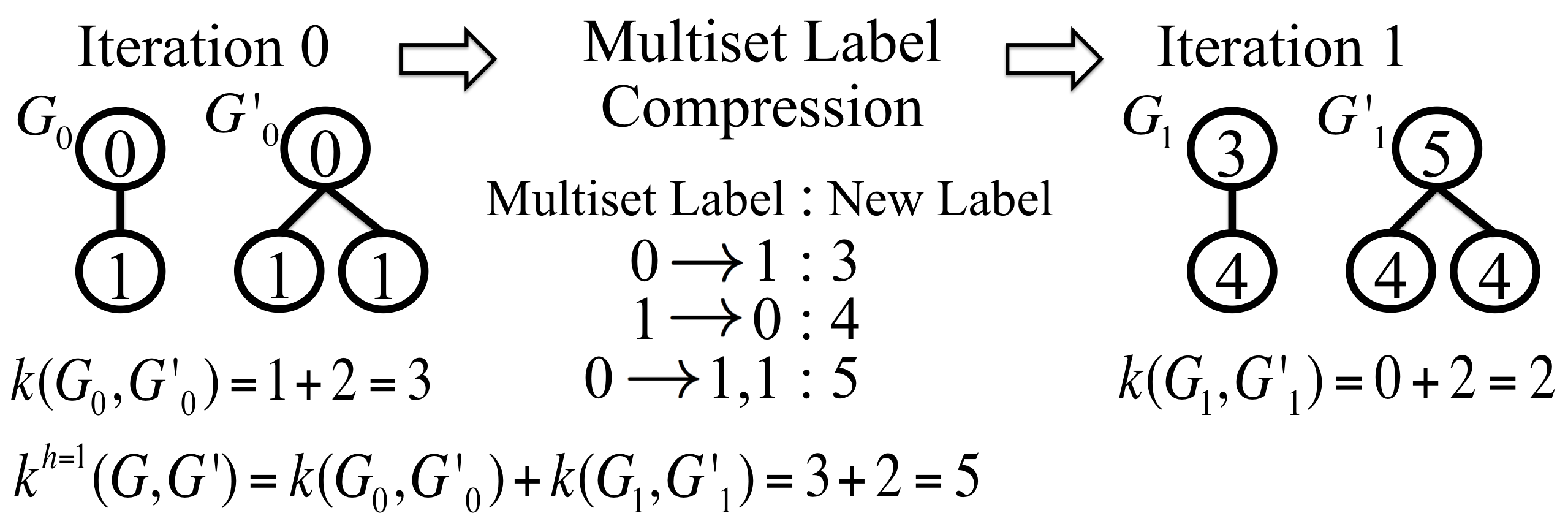}
\vspace{.08in}
\caption{\small{Toy example of the WL graph kernel. In a pre-processing step, raw node labels are converted to indices to facilitate kernel computation.}}
\label{fig:toy}
\end{figure}

Figure~\ref{fig:toy} illustrates how to calculate the WL graph kernel between graphs \begin{small}$G$\end{small} and \begin{small}$G'$\end{small} for degrees of neighbor up to 1 ($h$=1). Iteration $i$=0 for degree of neighbor 0 (stepsize 0) compares only the nodes of the original graphs. Nodes with label \begin{small}\textit{0}\end{small} have one match; nodes with label \begin{small}\textit{1}\end{small} have two matches. This gives a total similarity of three. The neighborhoods for each node are then augmented to compute similarity when iteration $i$=1, which compares the nodes and their first degree neighbors. New labels (i.e. \begin{small}\textit{3}\end{small}, \begin{small}\textit{4}\end{small}, and \begin{small}\textit{5}\end{small}) are assigned to represent each node and its first degree neighbors, and similarity of the relabeled graphs is 2. Therefore, \begin{small}$k^{h=1}(G,G')=k(G_{0},G'_{0})+k(G_{1},G'_{1})=3+2=5$\end{small}.

\subsection{Node Edge Weighting Graph Kernel}
\label{sec:new}

The WL kernel is efficient at neighborhood augmentation but there is no distinction between different node types, and node augmentation gathers up all nodes for a given degree. The 1-degree WL feature for the Designated Entity (DE) node in Figure \ref{fig:semgraph_all} is \begin{small}$<$DE$\rightarrow$Spkr,Msg,Msg$>$\end{small}, i.e. DE fills a \textsc{Speaker} and two \textsc{Message} elements (one for the \textit{Statement} frame and the other for the \textit{Convey\_importance} frame). No credit for partial matching is given when this graph instance is compared to another instance where DE just fills the \textsc{Message} element of the \textit{Convey\_importance} frame. To allow partial matching, and to take advantage of the type information of nodes and edges, we introduce a novel graph kernel: node edge weighting (NEW) graph kernel. 

Like WL, NEW also measures subgraph similarities through neighborhood augmentation. The kernel computation can be broken down into node kernels and edge kernels. Node and edge kernels are weighted Kronecker delta kernels ($\delta(\cdot,\cdot)$) that return whether the two objects being compared are identical. Define \begin{small}$w_{\mathcal{F}_{n}}$\end{small} for the weight of node \begin{small}$n$\end{small} of feaure type \begin{small}$\mathcal{F}$\end{small}, node label \begin{small}$\mathcal{L}$\end{small}, and \begin{small}$w_{<\mathcal{F}_{fr} \rightarrow \mathcal{F}_{to}>}$\end{small} for the weight of edge \begin{small}$e$\end{small} with from-node of feature type \begin{small}$\mathcal{F}_{fr}$\end{small} and to-node of feature type \begin{small}$\mathcal{F}_{to}$\end{small}. We have 

\begin{small}$k_{node}(n,n') = w_{\mathcal{F}_{n}} \cdot \delta(\mathcal{F}_{n},\mathcal{F}_{n'}) \cdot \delta(\mathcal{L}_{n},\mathcal{L}_{n'})$\end{small}, and
\begin{small}$k_{edge}(e,e') = w_{<\mathcal{F}_{fr} \rightarrow \mathcal{F}_{to}>} \cdot \delta(\mathcal{F}_{fr},\mathcal{F}_{fr'}) \cdot \delta(\mathcal{F}_{to},\mathcal{F}_{to'})$\end{small}. 
Define \begin{small}$k^{p}(G, G')$\end{small} to be the basis kernel for $p$-degree neighborhood; the kernel between graph \begin{small}$G$\end{small} and \begin{small}$G'$\end{small} is computed by recursion as in Equation~\ref{eq:newgk}. 

\vspace{-0.1in}
\begin{small}
\begin{equation}
\begin{split}
k^{p}(G, G')=\sum_{\text{all paths of length p } \in G,G'} k_{node}(n_{p}^{G},n_{p}^{G'}) \\
\prod_{i=1}^{p-1} k_{edge}(e_{i}^{G},e_{i}^{G'}) k_{node}(n_{i}^{G},n_{i}^{G'})
\label{eq:newgk}
\end{split}
\end{equation}
\end{small}
\vspace{-0.1in}

\noindent Dynamic programming can be used to improve the efficiency. Each entry in the dynamic programming table is a tuple of \begin{small}$<$$G,G',n_{i}^{G},n_{i}^{G'}$$>$\end{small}, where \begin{small}$n_{i}^{G}$\end{small} and \begin{small}$n_{i}^{G'}$\end{small} are nodes in graph \begin{small}$G$\end{small} and \begin{small}$G'$\end{small}.

\begin{figure}[t]
\centering{\includegraphics[width=2.5in]{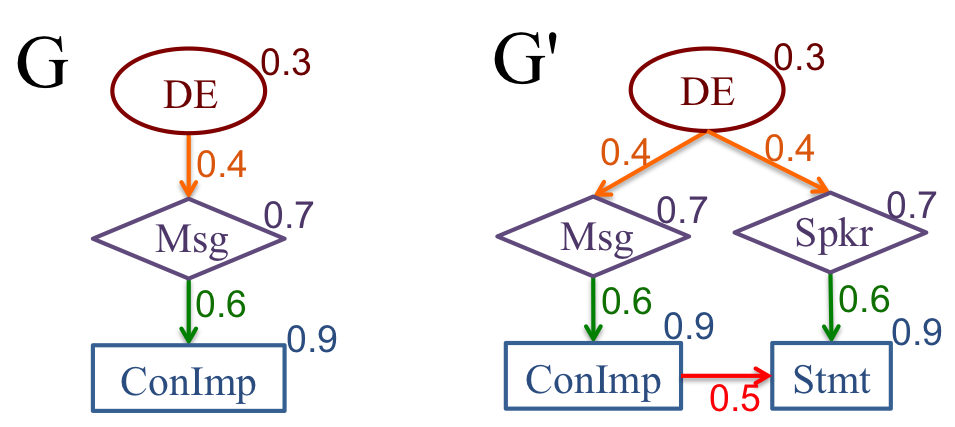}}
\caption{\small{Toy example of the \textsc{new} graph kernel.}}
\label{fig:newgk_toy}
\end{figure}

The toy example in Figure~\ref{fig:newgk_toy} illustrates how to calculate the \textsc{new} graph kernel between graphs \begin{small}$G$\end{small} and \begin{small}$G'$\end{small}. As with the \textsc{wl} kernel, \textsc{new} compares different degrees of node neighborhoods up to $p$ degrees of neighbors, and the final kernel is a sum of all basis kernels. For $p$=0, only the nodes of the original graphs are compared. Nodes with labels \begin{small}DE\end{small}, \begin{small}Msg\end{small} and \begin{small}ConImp\end{small} all have one match. With node weights as shown, \begin{small}$k^{p=0}$$=$0.3$+$0.7$+$0.9$=$1.9\end{small}. For \begin{small}$p$=1\end{small}, each node plus its one-degree neighbors are compared, and the relations between the nodes. Path \begin{small}DE $\rightarrow$ Msg\end{small} has a match. With node and edge weighting, \begin{small}$k^{p=1}$$=$0.3$*$0.4$*$0.7$=$0.084\end{small}. For the same reason, \begin{small}$k^{p=2}$$=$0.3$*$0.4$*$0.7$*$0.6$*$0.9$=$0.045\end{small}. There is no match for three degrees of neighbors, \begin{small}$k^{p=3}$$=$0\end{small}. 

Each basis kernel that corresponds to different neighborhood sizes are then normalized by the maximum of the evaluation between each graph and itself. For each graph kernel \begin{small}$k^{p}(G,G')$\end{small} we have a normalized 
\begin{small}$\hat{k}^{p}(G,G')$\end{small}:

\begin{small}
\begin{equation}
\hat{k}^{p}(G,G') = \frac{k^{p}(G,G')}{max(k^{p}(G,G),k^{p}(G',G')}
\end{equation}
\end{small}
\vspace{-0.1in}

This normalization ensures that a graph will always match itself with the highest value of 1 and other graphs with values between 0 and 1. 
The final kernel is an interpolation of basis kernels: 
\begin{small}
$k(G,G') = \sum_{p} \alpha_{p} \hat{k}^{p}(G,G')$
\end{small}
, where \begin{small}$\sum_{p} \alpha_{p}$=$1$\end{small}. 
Combining basis kernels is a common problem in machine learning and several multiple kernel learning techniques have been developed to allow benefits from multiple kernels \cite{Smits2002,Bach2004}.

\section{Financial News Analytics}
\label{sec:experiments}

We test the performance of OmniGraph with WL and NEW kernels on a polarity task: to predict the direction of price change for 321 companies from eight market sectors of Standard \& Poor's 500 index. On average, there are from 27 to 67 companies per sector. One of the biggest challenges of the financial domain is the unpredictability of the market. As noted above, use of NLP methods on news to predict price does well if it achieves better than random performance, as described in the Related Work. We rely on Student's T to test statistical significance of classification accuracy. We use the majority class label as a baseline, which ranges from 54\% to 56\%, depending on the market sector. Compared with three NLP benchmarks, only OmniGraph beats the baseline, and results are statistically significant. 

\subsection{Experimental Setup}
The experiments use Reuters news data from 2007 to 2013 for eight GICS\footnote{Global Industry Classification Standard.} sectors. Sentences that mention companies are extracted using high-precision, high-recall pattern matching on company name variants. A data instance for a company consists of an OmniGraph forest representing all the sentences that mention that company on a given day. On average, each data instance encodes from 4.11 to 7.18 sentences, and each company has an average total of from 605 to 858 sentences, depending on the sector. In work reported elsewhere, we found that we could expand the number of sentences per company using coreference by 15-30\%, depending on the sector. The additional sentences did not, however, improve performance (Anon). Sentences that mention companies by name tend to occur early in news articles, and are apparently more predictive.

A binary class label \{-1, +1\} indicates the direction of price change on the next day after the news associated to the data instance. The one-day delay of price response to news is due to \cite{Tetlock2008}. Only the instances with a price change of 2\% are included in our polarity prediction task.

Sentences are parsed using the MST dependency parser~\cite{mcdonald2005a}, which implements the Eisner algorithm~\cite{eisner1996} for dependency parsing, and provides an efficient and robust performance. For frame semantic parsing, we use SEMAFOR \cite{das11,das12naacl}, which generates state-of-the-art results on \textit{SemEval} benchmark datasets. 

For the learning, we found that no single stepsize performed best for a given company, much less the entire data set. We select the stepsize and weights of the basis kernels for NEW using grid search on 80\% of the data, where we use leave-one-out cross validation. The selected parameters for a given company are then used to test the average prediction performance on the 20\% of held-out data. 

\begin{table*}
\begin{center}
\begin{small}
\begin{tabular}{c r | c | c c c | l l}
GICS & Sector               & Baseline       & BOW            & DepTree        & SemTreeFWD     & OmniGraph$^{WL}$      & OmniGraph$^{NEW}$ \\ \hline
10 & Energy                 & 53.95$\pm$3.36 & 52.56$\pm$3.97 & 53.00$\pm$4.31 & 53.53$\pm$4.84 & 56.71$\pm$5.17$\ast$ & 56.90$\pm$4.20$\ast$ \\
15 & Materials              & 55.00$\pm$2.88 & 53.18$\pm$5.23 & 54.08$\pm$4.10 & 52.73$\pm$5.60 & 56.42$\pm$3.85$\ast$ & 56.49$\pm$3.26$\ast$ \\
20 & Industrials            & 54.25$\pm$3.85 & 52.89$\pm$5.91 & 53.10$\pm$3.49 & 52.90$\pm$5.21 & 55.29$\pm$4.44$\ast$ & 56.16$\pm$5.56$\ast$ \\
25 & Cons. Disc.            & 54.32$\pm$4.18 & 53.91$\pm$4.73 & 54.75$\pm$4.62 & 54.09$\pm$5.86 & 56.81$\pm$5.93$\ast$ & 57.49$\pm$5.63$\ast$ \\
30 & Cons. Staples          & 54.85$\pm$3.24 & 52.82$\pm$4.07 & 53.21$\pm$3.30 & 53.78$\pm$3.76 & 56.23$\pm$3.40$\ast$ & 60.64$\pm$10.7$\ast$ \\
35 & Healthcare             & 56.44$\pm$4.51 & 52.75$\pm$3.86 & 53.94$\pm$4.49 & 54.31$\pm$6.46 & 57.31$\pm$4.48 & 59.31$\pm$5.28$\ast$ \\
45 & IT                     & 53.95$\pm$4.07 & 52.42$\pm$3.64 & 52.80$\pm$4.07 & 52.79$\pm$6.84 & 55.38$\pm$4.92$\ast$ & 55.69$\pm$5.61$\ast$ \\
55 & Utilities              & 53.82$\pm$2.75 & 51.66$\pm$4.24 & 52.56$\pm$4.45 & 51.75$\pm$5.23 & 54.86$\pm$5.70 & 55.30$\pm$5.32 \\ \hline
\end{tabular}
\caption{\small{Mean accuracy by sector for the majority class baseline, three benchmarks, and OmniGraph with two graph kernels. The cases where the sector mean is significantly better than the baseline are marked by *. OmniGraph is significantly better than all three benchmarks in all cases.}}
\label{tab:results_sum}
\end{small}
\end{center}
\end{table*}

\begin{figure}[t]
\centering
    \includegraphics[width=0.4\textwidth]{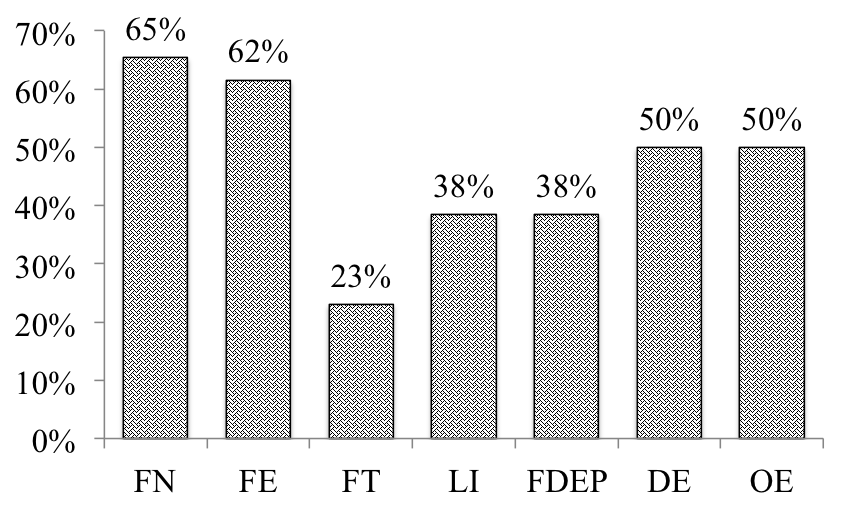}
    \caption{\small{OmniGraph$^{NEW}$ parameters for companies in Consumer Staples sector. It shows the total proportion across companies of node-edge weights for each feature type.}}
\label{tab:results_gics30}
\end{figure}

\subsection{Benchmark Methods}

Three benchmark methods are reported for comparison with OmniGraph:
(1) \textit{BOW}-a vector space model that contains unigrams, bigrams, and trigrams.
(2) \textit{DepTree}-a tree space representation based on the dependency parses used to create OmniGraph. The root is the sentence entry, and dependency relation types, such as SUB, OBJ, VMOD, and the lexical items, are represented as tree nodes. 
(3) \textit{SemTreeFWD}-a state-of-the-art representation for the price prediction task that is an enriched hybrid of vector and tree space~\cite{xie2013}. It includes semantic frames, lexical items, and part-of-speech-specific psycholinguistic features. Learning relies on Tree Kernel SVM~\cite{moschitti06}.\footnote{Data provided by the authors.}

\subsection{Features}

OmniGraph with \textsc{new} kernel learning shows a strong impact of stepsize and weighting of nodes and edges. In a detailed analysis of the 26 companies in GICS 30 (Consumer Staples), a sector with average amounts of news, all but three companies have non-zero coefficients on two or more of the basis kernels. Two of the three outliers rely only on stepsize 1 and the third on stepsize 2. Thirteen companies combine two basis kernels and the remaining ten combine three. On average, only 9\% of the features are non-relational (p=0). 
The other sectors have a similar trend. 

Grid search determines the stepsize, and also determines which node types to include during neighborhood augmentation; nodes are weighted $0$ or $1$. Figure~\ref{tab:results_gics30} shows the proportion of GICS 30 companies that use each of seven node types. The most important node types are frame names (FN) and frame elements (FE): more than 60\% of the companies need them to obtain the best performance. The next most frequent node types are designated entities (DE) and other entities (OE), each used by 50\% of companies. This result suggests that relations between companies are useful for price polarity. More than one third of the companies need the feature for dependencies between frames (FDEP), often involving complex sentences where multiples frames are evoked. The lexical item features (LI) have a contribution similar to FDEP. Note that depending on the stepsize, LI features from OmniGraph include lexical items (p=0), their dependencies and the frame elements they fill (p=1), and the frames to which the frame elements belong (p=2). Frame target (FT) is the least preferred feature.

\subsection{Results}

Table~\ref{tab:results_sum} summarizes the average accuracy for all eight sectors of the majority class baseline, the three benchmarks, and the two OmniGraph models. Both versions of OmniGraph significantly outperform the three benchmarks. The cells with asterisks represent a difference from the baseline that is statistically significant. OmniGraph$^{WL}$ beats the baseline with statistical significance in six sectors, and OmniGraph$^{NEW}$ in seven. Note that none of the benchmarks outperforms the baseline.

Despite the excellent performance of BOW for topical classification tasks, for this price prediction task it does poorly. Both DepTree and SemTreeFWD outperform BOW, which indicates that features derived from dependency syntax and semantic frame parsing improve performance. DepTree directly represents the dependency parse with both dependencies and words as nodes, without semantic information. The limitation of SemTree comes from its entity-centric representation -- the root node is the designated entity. The semantic frames without DE mentions are discarded, and a heterogeneous combination of trees and vectors are used for learning. Between \textsc{wl} and \textsc{new} learning on OmniGraph, \textsc{new} produces the best results. We suspect this is due to the high granularity of the features it generates, and its flexibility in assigning different weights to nodes and edges, depending on the node and edge feature types.

\section{GoodFor/BadFor Corpus}
\label{sec:experiments_gfbf}

To further test OmniGraph performance for entity driven text analytics, we used a recently introduced, publicly available dataset - the GoodFor/BadFor (\textit{gfbf}) Corpus\footnote{http://mpqa.cs.pitt.edu/corpora/gfbf/} \cite{Deng2013}, which is part of MPQA~\cite{Wiebe2005}. \textit{gfbf} has been annotated for two fine-grained sentiment judgments: 1) benefactive/malefactive event annotation, and 2) writer attitude. The benefactive/malefactive task asked annotators to identify the affected entity (the object) and the entity causing the event (the agent), and to label whether the agent and the event is benefactive or malefactive on the object. We treat the object as the designated entity. The writer attitude task asked annotators to identify the writer's attitude towards the agent and the object. We treat both the agent and the object as the designated entity in turn. 

\begin{table}
\begin{center}
\begin{small}
\begin{tabular}{l | c | c}
                  & Benef/Malef & WriterAttitude \\ \hline \hline
Baseline          & 56.65 & 55.61 \\ \hline \hline
BOW               & 67.13$\pm$2.68 & 66.61$\pm$1.90 \\
DepTree           & 72.10$\pm$2.41 & 66.16$\pm$1.76 \\
SemTreeFWD        & 72.51$\pm$2.22 & 65.32$\pm$2.05 \\ \hline
OmniGraph$^{WL}$  & \textbf{83.17$\pm$1.93} & \textbf{73.10$\pm$1.64} \\
OmniGraph$^{NEW}$ & \textbf{82.42$\pm$2.04} & \textbf{74.24$\pm$1.58} \\
\hline
\end{tabular}
\caption{\small{Mean accuracy for \textit{gfbf} data experiments, where OmniGraph (in boldface) significantly outperforms the baseline and three benchmarks.}}
\label{tab:results_gfbf}
\end{small}
\end{center}
\end{table}

We use the percentage of the majority class as the baseline, and compare the same five methods as in our previous experiment. Table~\ref{tab:results_gfbf} summarizes the results. On the benefactive/malefactive task, BOW obtains a 10\% improvement over the baseline. Structured representations significantly improve over BOW. SemTreeFWD, which incorporates the semantic frame features and a sentiment lexicon, improves the performance by another 5\%. The dependency tree performance is similar to SemTreeFWD. OmniGraph with graph kernel learning (WL or NEW kernels) performs much better. The writer attitude task is a more difficult one with a slightly lower baseline, and a much lower inter-annotator agreement~\cite{Deng2013}. Dependency trees and semantic trees do not improve over BOW. Both versions of OmniGraph, however, have superior performance.

\begin{figure*}[t]
\centering
{\footnotesize
\begin{tabular}{m{0.18in} | m{1.5in} | m{0.75in} | m{2.8in} | m{0.3in}}\hline
 & \centering{Graph Features} & \centering{Feature Types} & \centering{Example Sentences} & Label \\ \hline\hline
1 & \includegraphics[width=1.5in]{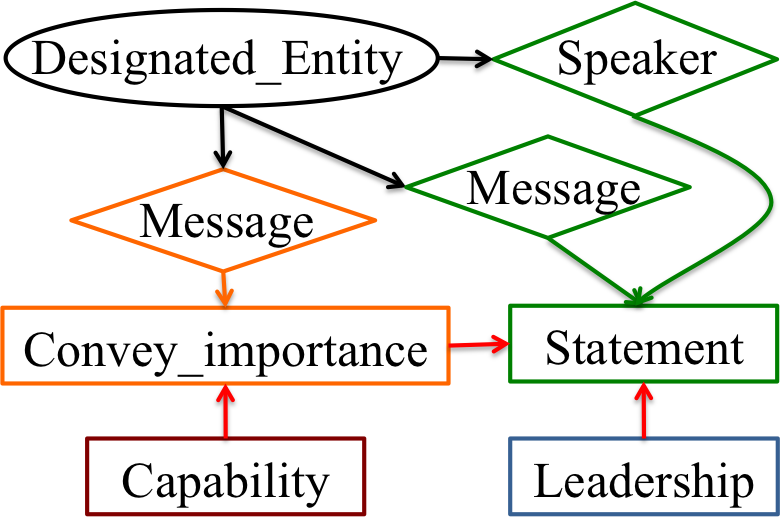} &  Frame names, \newline Frame \newline elements, \newline Dependencies \newline among frames & 
``This milestone \textcolor{Bittersweet}{highlighted} the \textbf{Boeing} KC-767's \textcolor{red}{ability} to perform refueling operations under all lighting conditions," \textcolor{OliveGreen}{said} George Hildebrand, \textbf{Boeing} KC-767 Japan \textcolor{blue}{program manager}. \newline
``This benchmark \textcolor{Bittersweet}{underlines} how \textbf{Intel} \textcolor{red}{can} collaborate to innovate and drive real performance and total cost of ownership benefits for our clients," \textcolor{OliveGreen}{said} Nigel Woodward, \textcolor{blue}{Global Director}, Financial Services for \textbf{Intel}.
&
\includegraphics[width=0.25in]{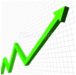} \newline GICS20 \newline GICS35 \newline GICS45 \\\hline
2 & \includegraphics[width=1.5in]{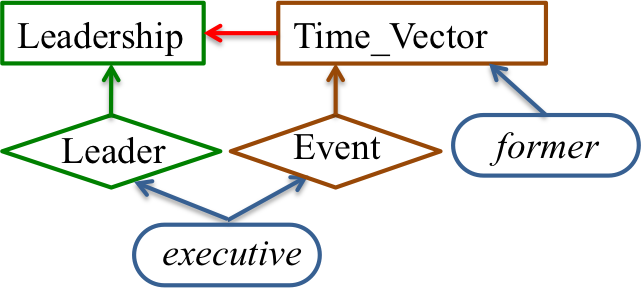} & Frame names, \newline Frame \newline elements, \newline Frame target, \newline Lexical items & 
The \textcolor{blue}{former} \textbf{Yahoo} \textcolor{orange}{executive} is a heavyweight in the online search and advertising area, with 20 U.S. patents. \newline
Picateers has hired Dan Levin, a \textcolor{blue}{former} senior \textcolor{orange}{executive} with \textbf{Intuit Inc}, to serve as interim CEO until a permanent leader is hired.
&
\includegraphics[width=0.25in]{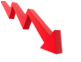} \newline GICS45 \\ \hline
3 & \includegraphics[width=1.3in]{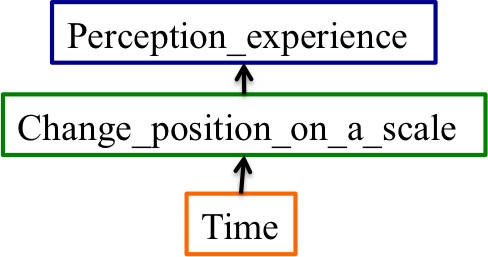} &  Frame names, \newline Dependencies among frames & 
\textbf{Wyndham} \textcolor{red}{have seen} profits \textcolor{blue}{soar} \textcolor{orange}{in recent years} as robust demand has allowed them to steadily raise rates. \newline
\textbf{Family Dollar} and \textbf{Walmart} are also expected to \textcolor{red}{see} same store sales \textcolor{blue}{growth} \textcolor{orange}{over the next 60 days}.
&
\includegraphics[width=0.25in]{up.png} \newline GICS25 \newline GICS30 \\ \hline
4 & \includegraphics[width=1.5in]{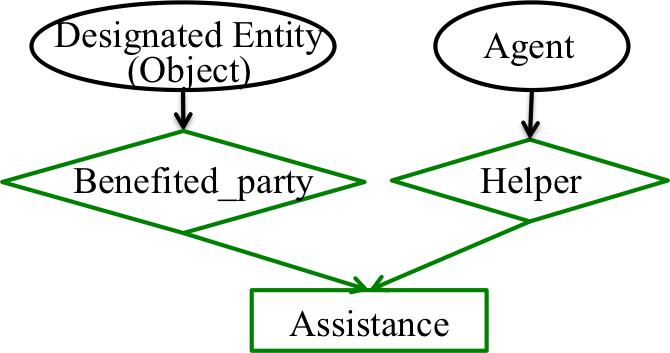} & Entities, \newline Frame name, \newline Frame \newline elements & 
Looking ahead to the benefits health care reform will bring in future years, the law also established \textbf{[a pregnancy assistance fund$_{Agent}$]} that will provide \$250 million over the next decade to help \textbf{[pregnant and parenting women and teens with services for those victimized by domestic or sexual violence$_{Object}$]}.
&
\includegraphics[width=0.25in]{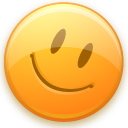} \\\hline
5\&6 & 
	\begin{subfigure}[b]{0.7in}
                \includegraphics[width=0.7in]{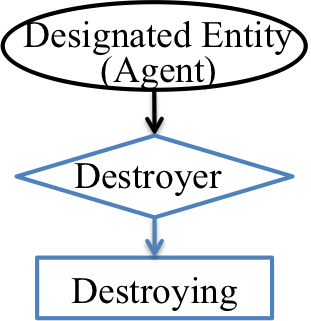}
        \end{subfigure}%
        ~~~~
        \begin{subfigure}[b]{0.7in}
                \includegraphics[width=0.7in]{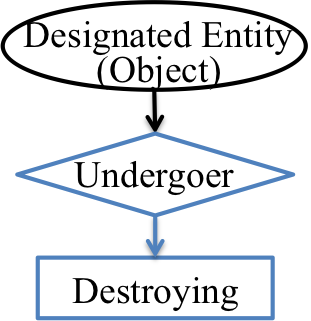}
        \end{subfigure}
 & Entities, \newline Frame name, \newline Frame \newline elements & 
By utilizing peer review practices which would not stand muster under standard constitutional law, \textbf{[hospital and health systems$_{Agent}$]} can label anyone a disruptive, unruly or uncooperative physician and destroy \textbf{[their ability to work$_{Object}$]}.
&
\includegraphics[width=0.25in]{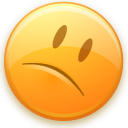} \newline \newline \includegraphics[width=0.25in]{happy.png} \\\hline
\end{tabular}
\caption{\small{Sample OmniGraph features for the financial news analytics and the GoodFor/BadFor task.}}
\label{fig:sample_features}
}
\end{figure*}

\section{Discussion}

OmniGraph with graph kernel learning exhibits superior performance over vector and tree space models in both experiments. To understand what contributes to the predictive power of OmniGraph models, we use mutual information to rank features discovered by OmniGraph. Compared to the vector and tree representations, the graph-structured features are more expressive, and can be interpreted. Figure~\ref{fig:sample_features} presents six highly ranked features from our experiments. Features 1-3 are from the financial news analytics task. Feature 1 is a complex feature with frame names, frame elements, and the dependencies among frames. It generalizes over multiple sectors and predicts a positive change in price. It is the feature that corresponds to the example sentences in the Introduction. Feature 2 combines frame names, frame elements, a frame target, and two lexical items to capture an interesting pattern: referring to the former leader of a company predicts a negative price change. Feature 3 is a 2-degree neighbor subgraph that consists of three frames and their inter-dependencies. This feature represents that the designated entity experiences a change over a time period. The feature generalizes across many different wordings, and although the feature does not directly encode direction of change, it happens that this feature rarely occurs in a negative description. It is therefore an example of a positive sentiment feature that is detected without reliance on a sentiment lexicon or on explicit polarity information.

Features 4-6 are from the \textit{gfbf} experiment. Feature 4 is a top ranked predictor for the benefective/malefactive task, and it predicts a positive affect toward the object. It captures the relation between the \textit{Agent} and the \textit{Object} in an \textit{Assistance} scenario where the \textit{Agent} fills the \textsc{Helper} role and the affected \textit{Object} fills the \textsc{Benefited\_party} role. Row 5 contains two features that are predictive in the writer attitude task. Recall that a writer can have different attitudes towards the \textit{Agent} and the \textit{Object}. Our approach is able to distinguish different roles of different entities of interest for the same sentence, and make separate predictions. 

As seen above, OmniGraph is very good at modeling complex intra-sentence semantic relations. Inspired by the work of \newcite{Galitsky2014160}, who constructed dependency parse forests for paragraphs of text, one of our future directions is to extend OmniGraph to incorporate discourse information. An obvious choice would be to encode inter-sentential discourse relations as one or more new edge types to connect the OmniGraphs that correspond to distinct sentences.

\section{Conclusion}

In this study, we have presented a novel graph-based representation -- OmniGraph -- with \textit{Weisfeiler-Lehman} and \textit{node edge weighting} graph kernel learning, for entity-driven semantic analysis of documents. This method exhibits superior performance in a text-forecasting task that uses financial news to predict the stock market performance of company mentions, and a fine-grained sentiment task. OmniGraph's advantages stem from the use of semantic frames to generalize word meanings in a flexible and extensible graph structure, where rich relational linguistic information, such as dependencies among frames and lexical items, can be modeled and learned with graph kernels that make feature engineering part of the learning. The resulting graph features are able to reflect deeper semantic patterns beyond words, and to help provide insights into the problem domain. Here, we applied OmniGraph to two rather distinct problems to illustrate that it could potentially support a wide range of NLP classification problems. On top of OmniGraph's capability of modeling complex intra-sentence semantic relations, a future direction is to model inter-sentence relations through discourse structure to form a more linguistically informed document-level representation.

\end{document}